\pdfoutput=1

\documentclass[11pt,a4paper]{article}
\usepackage[hyperref]{emnlp2020}
\usepackage{times}
\usepackage{textcomp}

\usepackage{latexsym}

\usepackage{microtype}

\usepackage{multirow}
\usepackage{tabularx}
\usepackage{booktabs}
\usepackage{amsmath}
\usepackage{placeins}
\usepackage{graphicx}
\aclfinalcopy %
\def\aclpaperid{157} %

\title{Incorporating Terminology Constraints in Automatic Post-Editing}

\author{David Wan$^{1}$,
    Chris Kedzie$^{1}$,
    Faisal Ladhak$^{1}$,
    Marine Carpuat$^{2}$
    and Kathleen McKeown$^{1}$
\\
  $^{1}$ Columbia University,
  $^{2}$ University of Maryland\\
  dw2735@columbia.edu, 
  \{kedzie,faisal,kathy\}@cs.columbia.edu,
  marine@cs.umd.edu\\
  }

\date{}

\begin{document}
\maketitle

\begin{abstract}

Users of machine translation (MT) may want to ensure the use of specific lexical terminologies. While there exist techniques for incorporating terminology constraints during inference for MT, current APE approaches cannot ensure that they will appear in the final translation. In this paper, we present both autoregressive and non-autoregressive models for lexically constrained APE, demonstrating that our approach enables preservation of 95\% of the terminologies and also improves translation quality on English-German benchmarks. Even when applied to lexically constrained MT output, our approach is able to improve preservation of the terminologies. 
However, we show that our models do not learn to copy constraints systematically and suggest a simple data augmentation technique that leads to improved performance and robustness.

\end{abstract}

\newcommand{\srctoks}{\mathbf{x}}
\newcommand{\srctok}{x}
\newcommand{\srcsize}{m}
\newcommand{\mttoks}{\boldsymbol{\gamma}}
\newcommand{\mttok}{\gamma}
\newcommand{\mtsize}{n}

\newcommand{\tgttoks}{\mathbf{y}}
\newcommand{\tgttok}{y}
\newcommand{\tgtsize}{o}
\newcommand{\srclang}{\mathcal{S}}
\newcommand{\tgtlang}{\mathcal{T}}
\newcommand{\numcons}{t}
\newcommand{\srccon}{\check{x}}
\newcommand{\tgtcon}{\check{y}}
\newcommand{\cons}{\mathcal{C}}
\newcommand{\srccons}{{\boldsymbol{\check{x}}}}
\newcommand{\tgtcons}{{\boldsymbol{\check{y}}}}
\newcommand{\srcconsize}{j}
\newcommand{\tgtconsize}{k}
\newcommand{\apptoks}{\mathbf{x}^+}
\newcommand{\reptoks}{\mathbf{x}^-}
\newcommand{\appfacs}{\mathbf{s}^+}
\newcommand{\repfacs}{\mathbf{s}^-}

\section{Introduction}

Automatic post-editing (APE) aims to improve the quality of the output of an arbitrary machine translation (MT) system by pruning systematic errors and adapting to a domain-specific style and vocabulary \cite{simard-etal-2007-statistical, chatterjee-etal-2018-findings}. %
Although previous work has shown the usefulness of APE to prune errors by focusing on improving the translation error rate (TER), few have studied the 
effect of 
incorporating lexical constraints. 

There are several use cases where such a system would be beneficial. For example, content providers meticulously curate lists of terminologies for their domains that indicate preferred translations for technical terms.
Lexically constrained APE would also be useful for cross-lingual information retrieval. When displaying snippets from retrieved documents, the query term should  appear in the translation output (if it does in the source) as it can make relevance clear to the end user. Here, the query serves as the term.

While recent approaches allow inference time adaptation of NMT systems using these terminologies \cite{lexical-nmt, post-vilar-2018-fast}, post-editing  translations with a generic APE system may lead to dropped %
terms. A constraint-aware APE system would allow to fix systematic translation errors, while keeping the terminologies intact.

Inspired by \citet{lexical-nmt}, we consider a range of representations which augment input sequences with constraint tokens and factors
for use in an autoregressive Transformer (AT) APE model.
Using this approach, the constraints are explicitly represented in the encoder input sequence, and the model learns to prefer translations that contain the supplied terminologies during decoding. We also explore the use of the Levenshtein Transformer (LevT) \cite{levt}, a 
non-autoregressive Transformer (NAT) model. The LevT model applies neatly to the APE task since the decoder can be initialized with an incomplete sequence to be refined. Additionally, multiple corrections can be 
made simultaneously, yielding a decoding speedup over autoregressive models.

We then show that constrained APE improves translation quality and terminology preservation on top of both unconstrained and constrained MT.
While both constrained and unconstrained APE models perform similarly on reducing systemic errors in the MT output, they differ in their ability to preserve %
terminology constraints.
When applying unconstrained APE on top of constrained MT, we find a 12.6\% relative drop of supplied terminology constraints as compared to a fully constrained MT to APE pipeline.

We experiment extensively with variations of both AT and LevT models, testing 
on both PBMT and NMT English to German WMT APE tasks \cite{chatterjee-etal-2018-findings}. 
Under all scenarios, the model performs post-editing while satisfying terminology constraints when supplied.

Our evaluation of both constrained AT and NAT models on PBMT and NMT APE benchmarks shows that
both models correctly translate more than 95\% of terminology constraints, with the NAT model achieving the highest coverage of terminologies at the expense of post-editing quality.

Finally, using constraints constructed with synonyms and antonyms, we show that our models do not learn to copy constraints systematically, and introduce a simple data augmentation strategy to improve the preservation of unusual constraints.

\begin{figure*}[ht]
    \centering
    \begin{tabularx}{\linewidth}{ r X }
    \toprule
        Source $(\srctoks)$  & The Gradient tool also provides most of the same \textbf{features} as the Gradient panel .\\
    \midrule
        Append $(\apptoks)$ & The$_{0}$ Gradient$_{0}$ tool$_{0}$ also$_{0}$ provides$_{0}$ most$_{0}$ of$_{0}$ the$_{0}$ same$_{0}$ \textbf{features$_1$} \textbf{Funktionen$_2$} as$_{0}$ the$_{0}$ Gradient$_{0}$ panel$_{0}$ .$_{0}$ \\
        Replace $(\reptoks)$ & The$_{0}$ Gradient$_{0}$ tool$_{0}$ also$_{0}$ provides$_{0}$ most$_{0}$ of$_{0}$ the$_{0}$ same$_{0}$ \textbf{Funktionen$_2$} 
        as$_{0}$ the$_{0}$ Gradient$_{0}$ panel$_{0}$ .$_{0}$ \\
    \midrule
        MT $(\mttoks)$ & Das$_{3}$ Verlaufswerkzeug$_{3}$ bietet$_{3}$ außerdem$_{3}$ die$_{3}$ meisten$_{3}$ der$_{3}$ gleichen$_{3}$ Merkmale$_{3}$ wie$_{3}$ das$_{3}$ Verlaufsbedienfeld$_{3}$ .$_{3}$ \\
    \midrule
       Post-Edit $(\tgttoks)$ & Das Verlaufswerkzeug bietet fast dieselben \textbf{Funktionen} wie das Verlaufsbedienfeld . \\
    \bottomrule
    \end{tabularx}
    \caption{An example of the inputs and output for the constrained APE task. Source is the source sentence. Post-Edit is the corrected MT sentence. We show the \textit{Append} and \textit{Replace} method to incorporate terminologies on the source side. Factors indicated for each word as source word (0), source constraint (1), target constraint (2), and MT word (3). The terminology pair for this example is (\textit{features}, \textit{Funktionen}). }
    \label{example_input}

\end{figure*}

To summarize, our contributions are as follows:
\begin{enumerate}
    \item We propose the terminology constrained APE task and evaluate several AT and LevT model variants for incorporating lexical constraints.
    
    \item We empirically show that constrained APE is necessary to preserve terminology constraints in a MT to APE pipeline.
    \item We analyze the robustness of the constraint translation behavior and suggest a simple data augmentation technique that both improves translation quality and increases
    the number of correctly translated %
    terms.
\end{enumerate}

\section{Related Work}
\subsection{MT with Terminology Constraints}
Integrating terminology constraints into translation can be divided into two approaches: constrained decoding and input sequence modification.

Constrained decoding modifies the decoding process to enforce the generation of the specified terminologies. This includes methods that modify beam search, such as grid beam search \cite{hokamp-liu:2017:Long} and dynamic beam allocations \cite{post-vilar-2018-fast}. While these approaches are effective in including terminologies,
they come with an increase in inference time due to 
the added overhead in the search algorithm.

The LevT \cite{levt}, which uses
a  non-autoregressive decoding procedure, 
can initialize its decoder with a partial or incomplete output sequence.
By initializing the decoder output with terminology constraints, \citet{lexical-levt} train a LevT
model to perform constrained decoding. Unlike constrained search methods in autoregressive models, this initialization technique does not add any significant overhead to 
the decoding process. 
When modified to disallow deletion of %
terms and insertion between consecutive terminology tokens, LevT is able to retain all terminologies without affecting the performance and speed.

Alternatively, \citet{lexical-nmt} propose modifying the encoder input sequence to represent terminology constraints. During training, the model learns to identify constraints in the input sequence, and
translate them appropriately during decoding. 
This approach has the benefit of not adding additional overhead during inference.

\subsection{Automatic Post-Editing}
The APE task has gone through many iterations, since it was originally proposed by \newcite{simard-etal-2007-statistical}. Initially, the task was to improve an unknown phrase-based machine transition (PBMT) system.
An additional task to fix errors of an NMT system was introduced at WMT 2018 \cite{chatterjee-etal-2018-findings}.

For the APE tasks,
the use of the multi-source variant of the neural encoder-decoder model is the most popular approach \cite{bojar-etal-2017-findings}, with the Multi-source Transformer (MST) instantiation \cite{2018-ape-pbmt-sota} achieving state-of-the-art results in 2018.
Based on the AT model \cite{NIPS2017_7181}, the MST model consists of two Transformer encoders and a single decoder. The source sentence %
and the MT system output 
are fed separately to the two encoders, where the outputs are concatenated and then fed into the decoder to perform post-editing.

Recent work has explored alternative architectures for APE. The winner of 2019 APE tasks \cite{2019-ape-nmt-sota}, for example, uses
a BERT-based encoder and decoder.
\citet{levt} both introduce the LevT model and demonstrate its utility on an APE task.

\section{Constrained APE}

The task of APE is to correct systematic errors in an MT system output. An APE model takes as input two sequences:
the source language sentence to be translated and the
translation of this sentence into the target language by
an MT system. The intended output is a corrected version
of the MT system's initial translation \cite{simard-etal-2007-statistical}.

Constrained APE allows for the specification of terminology
constraints: a translation for one or more phrases
in the source language input may be pre-specified as additional input. The constrained APE model must use the supplied terminology constraints when performing the APE task. 

Formally, let $\srctoks = \left[ \srctok_1, \ldots, \srctok_\srcsize \right]$ and  $\mttoks=\left[\mttok_1,\ldots,\mttok_\mtsize \right]$ be the source language sentence and the initial MT translation respectively. The tokens $\srctok_i$ and $\mttok_i$ are drawn from
the source and target language vocabularies $\srclang$ and $\tgtlang$ respectively. The target
post-edited sentence is a sequence $\tgttoks = \left[ \tgttok_1,\ldots,\tgttok_\tgtsize \right]$, with tokens $\tgttok_i$ also drawn from $\tgtlang$. 

We are also given a series of $\numcons$ translation constraints, 
$\cons = \left\{\left(\srccons^{(1)}, \tgtcons^{(1)}, \right),\ldots, \left(\srccons^{(\numcons)}, \tgtcons^{(\numcons)}, \right) \right\}$,
where each constraint $\left(\srccons^{(i)}, \tgtcons^{(i)} \right) \in \srclang^* \times \tgtlang^*$ is a tuple of source language phrase, $\srccons^{(i)} = \left[\srccon_1,\ldots, \srccon_\srcconsize \right],$ and its desired
translation, $\tgtcons^{(i)} = \left[\tgtcon_1,\ldots,\tgtcon_\tgtconsize \right]$, into the target language.

The goal of the constrained APE task is
to learn a mapping of $\srctoks,\mttoks,$ and $\cons$ to the target post-edited translation $\tgttoks$. Crucially, when a source side
constraint $\srccons^{(i)}$ matches a sub-sequence 
in $\srctoks$, it is required that the sub-sequence be translated as $\tgtcons^{(i)}$.
See \autoref{example_input} for an example.

\section{Models}

While there are existing models to address the APE task, and the lexical constrained MT task, it is not clear how to represent lexical constraints for APE models which, unlike MT models, take two sequences as input. We propose several techniques to incorporate constraints as additional inputs to the APE encoder by combining the input sequence modification
used in constrained MT \cite{lexical-nmt}
with the MST method of \citep{tebbifakhr-etal-2018-multi}.
For decoding,
we experiment with both the AT and the LevT decoders. The LevT decoder can additionally take advantage of different decoder initialization strategies for constrained decoding.

We first briefly show how we encode terminology constraints in the input sequence, before describing how they are incorporated into the MST and LevT APE models
specifically.

\begin{figure}[t]
    \centering
    
        \begin{tabular}{r c c}
        \toprule
        Model & Input & Init. \\
        \midrule
        MST & $\srctoks, \mttoks$ & -- \\
        MST Append  & $\apptoks, \mttoks$ & -- \\
        MST Replace & $\reptoks, \mttoks$ & -- \\
        \midrule
        LevT & $\srctoks$ & $\mttoks$ \\
        LevT Append & $\apptoks$ & $\mttoks$ \\
        LevT Replace & $\reptoks$ & $\mttoks$\\
       MS LevT & $\srctoks, \mttoks$ & $\tgtcons^{1}, \ldots, \tgtcons^{(\numcons)}$ \\
        \bottomrule
    \end{tabular}
    \caption{Setup for the models by the input and initialization at inference. %
    }
    \label{model_setup}
\end{figure}

\subsection{Encoding Lexical Constraints for APE in the Input Sequence}
\label{sec:input_mod}

In the APE setting, the input to the model is
the source language sentence $\srctoks$ and its initial MT translation $\mttoks$. We also need to represent
in $\srctoks$ the translation constraints $\cons$.

For clarity, we describe the case of representing a 
single translation constraint $\left(\srccons, \tgtcons\right)$ where $\srccons = \left[\srccon_1, \ldots, \srccon_\srcconsize\right]$
is a source language constraint and $\tgtcons =
\left[\tgtcon_1, \ldots,
\tgtcon_\tgtconsize\right]$
is its target language translation. Our approach trivially generalizes to multiple constraints.
We represent the constraint $\left(\srccons, \tgtcons\right)$ in $\srctoks$ in one of two ways. Either by appending the target language constraint $\tgtcons$ after the occurrence of $\srccons$ in the input sequence,
or by replacing occurrences of $\srccons$ in $\srctoks$ with $\tgtcons$.

For example, if we had the constraint, $\left(\srccons,\tgtcons\right) = \left(\left[\srccon_1, \srccon_2\right], \left[\tgtcon_1\right] \right)$
and the source  input  $\srctoks = \left[ \srctok_1, \srctok_2, \srctok_3, \srctok_4\right]$, with $\left[\srccon_1, \srccon_2\right] = \left[\srctok_2, \srctok_3\right]$,
we would obtain the following input sequences for the append and replace methods:
\begin{itemize}
    \item \textit{(Append)}  $\apptoks = \left[\srctok_1, \srctok_2,\srctok_3, \tgtcon_1, \srctok_4\right]$
    \item \textit{(Replace)}  $\reptoks = \left[\srctok_1, \tgtcon_1, \srctok_4\right].$
\end{itemize}

To further differentiate the constraint terms from other tokens in the source sentence, a ``source factor'' is associated with each input token.
The source factor is equal to 1 or 2 to indicate a source or target side 
terminology constraint, while 0 indicates an 
unconstrained source token. For the above examples,
we would obtain the following source factors:
\begin{itemize}
    \item \textit{(Append)}  $\appfacs = \left[0,1,1, 2,0\right]$
    \item \textit{(Replace)}  $\repfacs = \left[0, 2, 0\right].$
\end{itemize}

The source input sequence and source factor sequence
are separately embedded and concatenated
before they are fed into the encoder.
See \autoref{example_input} for examples of the \textit{append} and \textit{replace} methods applied to a source sentence.

We now describe how we use these modified input sequences in the MST and LevT models. See \autoref{model_setup} for an overview of the the proposed models and their configuration.

\subsection{Multi-source Transformer}
The input to an APE model is a pair of sequences, the source sentence
and the MT output to be post-edited.
To accommodate these two sequences, we use
the MST model of \citet{tebbifakhr-etal-2018-multi}, which uses a separate 
Transformer to encode each sequence.
The outputs of each encoder are concatenated 
and attended to by the decoder.

We augment the encoder for the source sentence with the \textit{append} and \textit{replace} methods. %
\autoref{example_input} shows an example of the inputs for the \textit{append} and \textit{replace} methods, $\apptoks$ and $\reptoks$ respectively.  To account for the additional input of MT, $\mttoks$, for the source factors, we use 3 for each token in $\mttoks$. For Byte-Pair Encoding (BPE) \cite{sennrich-etal-2016-neural}, the corresponding source factor token is applied for all subword units.

We train three variants based on MST: an unconstrained version as the baseline (MST), and two constrained versions using the \textit{append} (MST Append) and \textit{replace} (MST Replace) methods as described in \autoref{sec:input_mod}.

\subsection{Levenshtein Transformer}
The LevT follows the Transformer encoder-decoder architecture. However, instead of a regular Transformer decoder, the model uses three consecutive layers to simulate the edit operations. The first layer predicts whether each token should be deleted or kept. The second layer predicts how many placeholder tokens to insert between every two consecutive tokens. The final layer then predicts the actual target token for each placeholder.

One benefit of using the LevT is its ability to initialize the decoding process with
an arbitrary sequence. The first iteration of the decoding process is typically initialized with $\mathbf{y^0} = [\textit{\textless s\textgreater}, \textit{\textless /s\textgreater}]$, but it is possible to initialize it with MT (i.e. $\mttoks$) and allow it to be %
subsequently refined.

Since LevT retains the single encoder and decoder structure, the changes to incorporate lexical constraints are straightforward; we apply the \textit{append} and \textit{replace} methods to the encoder input. 

We also try augmenting the LevT similarly to the MST. Here, we have two encoders for the source, $\srctoks$,  and MT, $\mttoks$, respectively. During inference, we initialize the decoding string with %
target-side constraint terms, $\tgtcons$, similar to the constrained decoding setup in \citet{lexical-levt}.

For multiple constraints, we sort the target side terms $\tgtcons^{(i)}$ by the order of the occurrence of $\srccons^{(i)}$ in the source $\srctoks$. When source and target word order diverge, we hope that the model will learn to reorder constraints correctly,
but leave experimentation with constraint ordering for future work.

We train four variants of the LevT model. An unconstrained baseline model (LevT), and two constrained variants, with the same architecture as the base LevT, that incorporate constraints in the source using the \textit{append} $(\apptoks)$  (LevT Append) and \textit{replace}  $(\reptoks)$ (LevT Replace) methods described in \autoref{sec:input_mod}. The decoder initialization for these models is the MT sentence, $\mttoks$, that needs to be edited. The final variant has a multi-source encoder, where $\srctoks$ and $\mttoks$ are fed into separate encoders. The decoder in this case is initialized with the target sequence of the terminology constraint(s).

\begin{table*}[ht]
\begin{center}
\begin{tabular}{c c  c c c  c c c }
\toprule
\multicolumn{2}{c}{\multirow{2}{*}{Dataset}} & \multicolumn{3}{c}{\# of Triplets } & \multirow{2}{*}{Term\%} & \multirow{2}{*}{TER} & \multirow{2}{*}{BLEU} \\
\cmidrule(lr){3-5}
& & Train & Valid & Test \\
\midrule
\multirow{3}{*}{PBMT} & artificial 4M & 4,390,180 & 1,000 & - & - & -  & - \\
& artificial 500K & \phantom{0,}526,368 & - & - & - & -  & - \\
& WMT'18 APE & \phantom{0,0}24,000 & 2,000 & 2,000 & 88.83 & 24.57 & 62.39 \\
\midrule
\multirow{2}{*}{NMT} & eSCAPE NMT & 4,999,102 & 1,000 & - & - & - & - \\
& WMT'19 APE & \phantom{0,0}13,442 & 1,000 & 1,023 & 89.52 & 16.92 & 74.60 \\
\bottomrule
\end{tabular}
\end{center}
\caption{\label{ape_dataset_stat} Statistics for data used. Term\%, TER, and BLEU are provided for do-nothing case of test set.}
\end{table*}

\section{Data}
\subsection{APE Datasets}
We use two standard English-to-German APE benchmark datasets, WMT18 PBMT \cite{chatterjee-etal-2018-findings} and WMT19 NMT \cite{chatterjee-EtAl:2019:WMT}. Both datasets are in the IT domain.
Each example from these datasets consists of three sequences:
(1) the source sentence $\srctoks$, (2) its MT output $\mttoks$, and (3) its post-edited target $\tgttoks$.

Since the official collections are relatively small, we augment them with
large synthetic datasets for pretraining: artificial \cite{junczysdowmunt-grundkiewicz:2016:WMT} and eSCAPE \cite{negri-etal-2018-escape}.
The artificial dataset is generated using round-trip translation of two PBMT systems. It is already cleaned and tokenized.
The eSCAPE dataset, containing 7,258,533 triplets, is created using NMT generated output from various parallel corpora. The data for eSCAPE is noisy, and we follow \citet{lee-etal-2019-transformer}'s procedure to filter the dataset, which results in around 5 million triplets. We then tokenize the filtered data using Moses \cite{koehn-etal-2007-moses}.\footnote{\url{www.statmt.org/moses/}}
For pretraining on the synthetic corpora, we set aside 1,000 randomly sampled triplets as our validation set.
\autoref{ape_dataset_stat} summarizes the statistics of both the evaluation and pretraining datasets.

For both tasks, we use the same preprocessing steps. After tokenization, we truecase the data using Moses. We then use BPE with 32,000 merge operations on the joined vocabulary of source and target language.

\subsection{Terminology Dataset}
\label{sec:termgen}

We create terminology sets for each APE dataset using Wiktionary.\footnote{We use the latest dump as of 06/18/2020}
We follow the procedure of \citet{lexical-nmt}, finding term translation pairs $(\srccons,\tgtcons)$ in Wiktionary such that $\srccons$ is present in the source sentence $\srctoks$ and $\tgtcons$ is present in the post-edited target sentence $\tgttoks$. We ignore stop words that appear on the source and target side.
To include more morphological variations, we include matches on stemmed versions of $\srccons$ and $\tgtcons$ using Snowball stemming.\footnote{\url{www.nltk.org/\_modules/nltk/stem/snowball.html}} We recover the unstemmed words from the pairs to be included in the terminology dataset. 
In order for the model to perform the APE task well when no constraints are supplied, we keep only 25\% of matched terminology constraints (i.e. we remove 75\% of constraints at random).

We split the terminology dataset into training and test sets so that terminology constraints provided at test time are not seen during training. We only use the training set for the training corpora of APE datasets, and use the test sets of the terminology on the validation and test set of the APE datasets. %
See \autoref{term_stat} for statistics of terminology coverage on the training, validation, and test splits.

With the given MT system, we can evaluate on terminology percentage for the do-nothing case, which is shown in \autoref{ape_dataset_stat}. The original MT model already achieves a high term percentage of around 90\% for PBMT and NMT tasks.

\begin{table*}[t]
\centering
\begin{tabular}{c c  c c c  c c c }
\toprule
\multicolumn{2}{c}{\multirow{2}{*}{Dataset}} & \multicolumn{3}{c}{\# of Triplets with Term. } & \multicolumn{3}{c}{Avg \# of Term. } \\
\cmidrule(lr){3-5}\cmidrule(lr){6-8}
& & Train & Valid & Test & Train & Valid & Test \\
\midrule
\multirow{3}{*}{PBMT} & artificial 4M & 1,605,075 & 345 & - & 1.25 & 1.25  & - \\
& artificial 500K & \phantom{0,}207,225 & - & - & 1.27 & -  & - \\
& WMT'18 APE & \phantom{0,00}6,037 & 834 & 528 & 1.15 & 1.24 & 1.34 \\
\midrule
\multirow{2}{*}{NMT} & escape NMT & 1,768,587 & 335 & - & 1.28 & 1.30 & - \\
& WMT'19 APE & \phantom{0,00}3,450 & 262 & 408 & 1.16 & 1.14 & 1.25 \\
\bottomrule
\end{tabular}
\caption{The number of training/validation/test instances that have at least one terminology constraint and the average number of terminology constraints for those instances.}
\label{term_stat}
\end{table*}

\section{Experiments}
We use the \textsc{fairseq} toolkit \cite{ott-etal-2019-fairseq} for implementing the MST and extending the LevT.\footnote{Our code is publicly available at \url{https://github.com/zerocstaker/constrained_ape}.}
We evaluate the models on translation error rate (TER) \cite{Snover06astudy} and BLEU \cite{papineni-etal-2002-bleu} using the official evaluation script\footnote{\url{www.dropbox.com/s/5jw5maariwey080/Evaluation_Script.tar.gz?dl=0}} for analyzing the post-editing performance. We also compute the percentage of target language term constraints present in the output (Term \%) to measure the performance of the constrained models. 

\subsection{Constrained MT-to-APE Cascades}
In our first experiment, we attempt to demonstrate the utility of constrained APE 
when applied to constrained MT. That is, we have some terminology constraints that
we want to preserve throughout the application of MT and subsequent APE. We conjecture that unconstrained APE applied on top of constrained MT will potentially
discard or re-translate previously translated constraints.

We experiment with all possible pipelines of MT to APE, i.e. the product of $\{\textrm{MT}, \textrm{Const. MT}\} \times \{\textrm{No APE}, \textrm{APE}, \textrm{Const. APE}\}$ with six total pipelines possible.

\paragraph{MT Models} To obtain MT models for this experiment we train both a constrained and unconstrained AT MT model using the default \textsc{fairseq} Transformer hyperparameters
and use the embedding size of 16 for the source factor embedding.
We follow the settings of \citet{lexical-nmt},
training an unconstrained transformer and a constrained model with append method to perform English-to-German translation using the Europarl and News Commentary data, and using the WMT 2013/2017 test set as validation and test set respectively.
The preprocessing steps follows that of the APE datasets.

For the constrained MT model we used the \textit{append} input modification method to make the model constraint aware. Terminology constraints are generated according to the method described in \autoref{sec:termgen}.
\citet{lexical-nmt} also released their Wiktionary terminology set (Wikt975)\footnote{\url{https://github.com/mtresearcher/terminology_dataset}} and we also show evaluation results using this terminology collection.
We report BLEU and terminology coverage (Term \%)
for our MT models on the WMT 2017 test set in 
\autoref{result_translation}.

\begin{table}[t]
\begin{center}
\begin{tabular}{c c c c c }
\toprule
& \multicolumn{2}{c}{Our Term.} & \multicolumn{2}{c}{Wikt 975} \\
\cmidrule(lr){2-3}\cmidrule(lr){4-5}
MT & Term\% & BLEU & Term\% & BLEU \\
\midrule
AT & 71.70 & 23.76  & 74.78 & 24.00 \\
AT App.  & 93.62 & 24.62 & 93.07 & 24.14 \\
\bottomrule
\end{tabular}
\end{center}
\caption{\label{result_translation} Translation result of vanilla and lexically constrained translation. }
\end{table}

\paragraph{APE Models} We use the MST and the append method as the unconstrained APE and constrained APE respectively.
We evaluate the MT to APE pipelines using the WMT'19 APE test set, replacing the provided MT in the triplet with the outputs from our unconstrained or constrained MT models.

We train the APE models using the eSCAPE corpus, where 1,000 triplets are used as validation set. We use the default \textsc{fairseq} Transformer hyper-parameters. For the constrained APE, we use the embedding size of 16 for the source factor tokens.

\subsection{Benchmark APE Tasks}

 The APE models are trained in two step fashion. First, a general APE system is trained using a synthetic dataset until convergence. Then the model is refined on the official dataset.
 For the PBMT task, we follow the training procedure of \citet{levt}. The model is pretrained on the artificial 4M dataset, and fine-tuned on the joined dataset of the 500K artificial dataset and the 10 times up-sampled official PBMT data. For the NMT task, we pretrain on eSCAPE and fine-tune on the official NMT data.

We use the default Transformer parameters for the MST variants, with an embedding size of 16 for source factors of the constrained APE models. 
For the LevT models, we follow the same setup and  hyper-parameters as described in \citet{levt}. 

\begin{table}[t]
\begin{center}
\begin{tabular}{c c c c}
\toprule
Pipeline & Term\%$\uparrow$ & TER$\downarrow$ & BLEU$\uparrow$ \\
 \midrule
\phantom{c}MT   & 45.33 & 70.78 & 15.28 \\
cMT  & 86.33 & 70.24 & 15.47 \\
\hline
\phantom{c}MT~~$\rightarrow$ \phantom{c}APE  & 55.35 & 59.56 & 22.87\\
cMT~~$\rightarrow$ \phantom{c}APE & 77.22 & 59.78 & 23.03\\
\phantom{c}MT~~$\rightarrow$ cAPE & 80.18 & \textbf{58.70} & \textbf{23.95}\\
cMT~~$\rightarrow$ cAPE & \textbf{88.38} & 59.77 & 23.08\\
\bottomrule
\end{tabular}
\end{center}
\caption{\label{result_usefulness} Result of different combinations of MT and APE systems. Constrained MT and APE are indicated cMT and cAPE respectively.}
\end{table}

We compare our models to the do-nothing case, where the output of the MT, $\mttoks$, is treated as the predicted post-edited sentence $\boldsymbol{\hat{\tgttoks}}$. The unconstrained variant also serve as a basis for comparing the performance of the constrainted APE models. We also compare our models to the winning system for the tasks, MS\_UEdin \cite{2018-ape-pbmt-sota} for PBMT 2018 and  Unbabel\_BERT \citet{2019-ape-nmt-sota} for NMT 2019.

\section{Results and Discussions}

\subsection{Constrained MT-to-APE Cascade } 

\autoref{result_usefulness} shows the result of the various combinations of MT and APE systems. 
Since the MT system is trained on news/parliamentary proceedings and not on the IT domain of the APE data, the translation quality is relatively low. 
Nevertheless, the constrained MT can include almost twice as many terminologies as the original model. Both APE systems improve the quality of the MT outputs, with constrained APE performing slightly better. However, constrained APE excels at including terminologies, as it consistently increases the terminology percentage from the previous MT output. 
When supplied with a constrained MT, the vanilla APE actually decreases the percentage of correct terminologies by 9\% (86.33\% to 77.22\%), whereas the constrained APE model can increase it by 2\% (86.33\%  to 88.38\%).

\subsection{Benchmark APE on PBMT Output}
\begin{table}[t]
\begin{center}
\begin{tabular}{l c c c c }
\toprule
 Models & Term\%$\uparrow$ & TER$\downarrow$ & BLEU$\uparrow$  \\
\midrule
Do-nothing & 88.48 & 24.25 & 62.99 \\
MS\_UEdin\ & 88.70 & \textbf{18.01} & \textbf{72.52} \\
\midrule
MST & 90.11 & 19.34 & 70.44 \\
MST Append  & 95.54 & 18.97 & 70.63 \\
MST Replace & 95.43 & 19.17 & 70.34 \\
\midrule
LevT  & 90.76 & 24.21 & 63.47 \\
LevT App. & 90.98 & 23.88 & 64.97 \\
LevT Rep.  & 91.41 & 23.94 & 64.96 \\
MS LevT & \textbf{97.50} & 20.39 & 68.57 \\
\bottomrule
\end{tabular}
\end{center}
\caption{\label{result_pbmt} Results for PBMT 2018. }
\end{table}
\autoref{result_pbmt} shows the results on the PBMT task. All MST variants improve from the do-nothing case, where the output is unchanged, i.e.$\mttoks = \boldsymbol{\hat{\tgttoks}}$. Using either the \textit{append} or \textit{replace} methods shows similar improvements in Term\%, increasing about 7\% points absolutely over the do nothing case. The terminology aware MST models also see small decreases in TER and small increases in BLEU relative to the unconstrained MST model. These results are encouraging as it shows that introducing terminology constraints does not interfere with the APE system's ability to fix systematic errors.

We were unable to reproduce the result by \citet{levt}; we see only small improvements with the LevT models relative to the do-nothing case.
Additionally, the \textit{append} and \textit{replace} variants yield only small increases in Term\% but are around 4-5\% points behind the equivalent MST model.
The MS LevT, however, achieves the highest terminology percentage of all models, while slightly underperforming the MST models on TER and BLEU.

None of our proposed models beat the SOTA baseline for this task on TER or
BLEU, but our best model on TER, MST Append,
is less than 1 percentage point worse in TER. At the same time, MST Append 
successfully translates 6.8\% more terminology constraints than the SOTA baseline.

\begin{table}[t]
\begin{center}
\begin{tabular}{l c c c c }
\toprule
models & Term\%$\uparrow$ & TER$\downarrow$ & BLEU$\uparrow$ \\
\midrule
Do-nothing & 90.22 & 16.84 & 74.73 \\
Unbabel\_BERT & 89.98  & \textbf{16.06} & \textbf{75.96}  \\
\midrule
MST & 90.66 & 16.46 & 75.61 \\
MST Append & 94.08 & 16.62 & 75.16 \\
MST Replace & 94.08 & 16.56 & 75.39 \\
\midrule
LevT & 90.41 & 17.28 & 74.17 \\
LevT App. & 91.59 & 17.32 & 74.25 \\
LevT Rep. & 90.61 & 17.14 & 74.46 \\
MS LevT & \textbf{98.04} & 17.71 & 73.64 \\
\bottomrule
\end{tabular}
\end{center}
\caption{\label{result_nmt} Results for NMT 2019.}
\end{table}

\subsection{Benchmark APE on NMT Output}

\begin{figure*}[ht]
    \centering
    \small
    \begin{tabularx}{\linewidth}{ c c X }
        \toprule
        \multirow{10}{*}{Original} 
        & Source $(\srctoks)$ & increasing the \textbf{magnification} can also make reshaping easier and more accurate . \\
         & MT $(\mttoks)$ & durch das Vergrößern der \textbf{Vergrößerung} können Sie außerdem das Umformen von Formen und präziser steuern .\\
         & Post-Edit $(\tgttoks)$ & durch das Vergrößern der \textbf{Vergrößerung} können Sie außerdem das Umformen von Formen präziser steuern . \\
         & MST Append & Durch das Vergrößern der \textbf{Vergrößerung} können Sie außerdem das Umformen von Formen erleichtern und präziser steuern . \\
         & MS LevT & Durch die zunehmende \textbf{Vergrößerung} können Sie außerdem das Umformen von Formen und präziser steuern . \\
        \midrule
        \multirow{6}{*}{Synonym} & Post-Edit $(\tgttoks)$ & durch das Vergrößern der \textbf{Magnifizierung} können Sie außerdem das Umformen von Formen präziser steuern .\\
         & MST Append & durch das Vergrößern der \textbf{Magnifizierung} können Sie außerdem das Umformen von Formen vereinfachen und präziser steuern . \\
         & MS LevT & eine Erhöhung der \textbf{Magnifizierung} kann außerdem das Umformen von Formen und präziser erleichtern . \\
        \midrule
        \multirow{6}{*}{Antonym} & Post-Edit $(\tgttoks)$ & durch das Vergrößern der \textbf{Verkleinerung} können Sie außerdem das Umformen von Formen präziser steuern . \\
         & MST Append & durch das Vergrößern der \textbf{Vergrößerung} können Sie außerdem das Umformen von Formen vereinfachen und präziser steuern . \\
         & MS LevT & durch die zunehmende \textbf{Verkleinerung} können Sie außerdem das Umformen von Formen und präziser steuern .\\
        \bottomrule
    \end{tabularx}
    \caption{Example of the outputs by the MST Append and MS LevT when a synonym and an antonym is supplied in place of the original terminology pair (\textit{magnification} - \textit{Vergrößerung}). The synonym \textit{Magnifizierung} (magnification) and antonym \textit{Verkleinerung} (diminishment) is used. }
    \label{example_copying}
\end{figure*}

\begin{figure*}[ht]
    \centering
    \small
    \begin{tabularx}{\linewidth}{ r  X }
        \toprule
        Source $(\srctoks)$ & if you use the Image Processor , you can \textbf{save} the files directly to JPEG format in the size that you want them .\\
        Post-Edit $(\tgttoks)$ & wenn Sie den Bildprozessor verwenden , können Sie die Dateien direkt im JPEG-Format in der gewünschten Größe \textbf{speichern} \\
        \midrule
        Synonym & wenn Sie den Bildprozessor verwenden , können Sie die Dateien direkt im JPEG-Format in der gewünschten Größe \textbf{sichern} \\
        Antonym & wenn Sie den Bildprozessor verwenden , können Sie die Dateien direkt im JPEG-Format in der gewünschten Größe \textbf{löschen} \\
        \bottomrule
    \end{tabularx}
    \caption{Example of data augmentation. The original  term pair is (\textit{save}, \textit{speichern}). We replace the target terminology \textit{speichern} with the synonym \textit{sichern} (to store for future use) or the antonym \textit{löschen} (to delete).}
    \label{example_data_augmentation}
\end{figure*}

\begin{table*}[t]
\begin{center}
\begin{tabular}{l c c c  c c c }
\hline
 & \multicolumn{3}{c}{WMT'19 APE} & \multicolumn{3}{c}{ Augmentation } \\
 \cmidrule(lr){2-4}  \cmidrule(lr){5-7}
 & Term\%$\uparrow$ & TER$\downarrow$ & BLEU$\uparrow$  & Term\%$\uparrow$ & TER$\downarrow$ & BLEU$\uparrow$ \\
\toprule
Do-nothing & 90.22 & 16.84 & 74.73 & 1.66 & 24.77 & 62.56 \\
\hline
MST Append & 94.08 & 16.62 & 75.16 & 7.47 & 24.92 & 61.80 \\
MST Append + \textit{pretrain} & 94.08 & 16.46 & 75.25 & 18.67 & 23.70 & 64.38\\
MST Append + \textit{pretrain} + \textit{ft} & 93.85 & \textbf{16.29} & \textbf{75.38} & 43.15 & \textbf{21.85} & \textbf{67.41}\\
\midrule
MS LevT & 98.04 & 17.71 & 73.64 & 43.57 & 33.07 & 54.33\\
MS LevT + \textit{pretrain} & \textbf{99.09} & 17.18 & 74.22 & 52.70 & 29.79 & 60.24\\
MS LevT + \textit{pretrain} + \textit{ft} & 98.41 & 17.00 & 74.66 & \textbf{63.07} & 29.66 & 60.47\\
\bottomrule
\end{tabular}
\end{center}
\caption{\label{result_augmentation} Results with data augmentation for the official APE data, as well on the augmented dataset consisting of synonyms and antonyms generated from Wiktionary. The size of the additional data for test set is 236. }
\end{table*}
\begin{table*}[t]
    \centering
   \begin{tabular}{l c c c c c c }
    \toprule
     & \multicolumn{3}{c}{Synonym} & \multicolumn{3}{c}{Antonyms } \\
     \cmidrule(lr){2-4}  \cmidrule(lr){5-7}
     & Term\%$\uparrow$ & TER$\downarrow$ & BLEU$\uparrow$  & Term\%$\uparrow$ & TER$\downarrow$ & BLEU$\uparrow$ \\
     \midrule
    MST Append & 7.33 & \textbf{1.31} & \textbf{97.80} & 8.88 & \textbf{1.06} & \textbf{98.01} \\
    MST Append + \textit{pretrain} & 16.75 & 2.81 & 94.19 & 28.88 & 3.81 & 92.94\\
    MST Append + \textit{pretrain} + \textit{ft}  &  38.74 & 5.48 & 88.86 & 66.66 & 6.46 & 87.50 \\
    \midrule
    MS LevT  & 41.36 & 19.57 & 70.60 & 57.77 & 17.56 & 74.25 \\
    MS LevT + \textit{pretrain} & 47.12 & 18.33 & 72.27 & 82.22 & 13.28 & 79.42 \\
    MS LevT + \textit{pretrain} + \textit{ft}  &  \textbf{43.97} & 18.25 & 73.25 & \textbf{77.78} & 11.99 & 80.41 \\
    \bottomrule
    \end{tabular}
    \caption{Structural change from the output of constrained models using the correct terminology. We split the dataset by synonyms and anotnyms, consisting of 191 and 45 samples respectively. }
    \label{result_syn_ant}
\end{table*}

\autoref{result_nmt} shows the result of the NMT task. This is a more difficult post editing task as the machine translated text from NMT systems is of a higher quality than PBMT systems, and the official training corpus is smaller than that of PBMT APE \cite{chatterjee-etal-2018-findings}. 
As further evidence of the difficulty of this task, the winning system of the WMT2019 APE shared task is able to achieve a mere 0.78 point decrease in TER.

The two terminology-aware MST models (\textit{append} and \textit{replace}) are able to improve Term\% over the baseline, at the cost of a slight increase in  TER and decrease in BLEU, but both are better than doing nothing.
The LevT and its variants perform worse than doing nothing in terms of TER and BLEU, but has a small gain in Term\%.
The MS LevT again achieves the highest Term\% but does
worse than the do-nothing case on TER and BLEU.

\section{Analyzing Constraint Translation Behavior}

Terminology constrained APE aims to add some degree of user control over the APE process without destabilizing the general post-editing behavior of the decoder.
However, the imposition of rare or unusual terminology constraints will
necessarily be in conflict with the decoder 
language model, which will give higher probabilities
to terminology translations found frequently in the training
data. 

In practice, a user may specify 
a terminology constraint that is not well represented
in the training distribution. For example, a user
may want a product description translated using
location specific brand names or marketing copy.
Ideally, a terminology constrained model would reliably produce these terms and use 
them appropriately even if they do not rank highly by the decoder. 

Additionally, it is desirable that the addition of terminology 
constraints does not lead to large changes in the model's output. Since terminology constraints 
are only bound to a word or phrase, the model should only need to make minimal changes between the unconstrained and constrained output. Large changes in output may make it harder for a user to anticipate
the effects of a constraint which may make 
constrained APE less useful in practice.

We refer to this behavior as
\textit{systematic copying}, i.e. the model should behave
in a transparent and stable way, enforcing 
terminology constraints even when they strongly disagree
with the decoder language model, while only making minimally
necessary changes in the output to do so.

By harvesting terminology constraints from the 
training data, we run the risk that the model simply learns to draw some translation hints from the supplied terminology constraints, but does not actually learn this systematic copying behavior.
That is, it never truly sees an out-of-sample
constraint that is extremely unlikely from the perspective of the decoder language model. 

To test whether our proposed models indeed learn this 
systematic copying behavior we perform a qualitative experiment,
comparing model outputs when supplying different constraints
for a source word, by varying whether the target language 
constraint was (a) the original target language constraint specified in the test set,
(b) a target language synonym of the original constraint term,
(c) a target language antonym of the original constraint term,
or (d) a totally random term in the target language. 

While antonym and random term constraints might not seem to correspond to realistic use cases, they let us to examine the effects of specifying a %
constraint where the source and target language terms are extremely
semantically divergent. Additionally, they simulate scenarios where translations for names vary dramatically by region. For example, the cleaning product called ``Mr. Clean'' in the U.S. is called ``Meister Proper'' in Germany.

As can be seen in \autoref{example_copying}, our qualitative exploration reveals that
synonym, antonym, and random terminology constraints are frequently 
not included in the output.  For example the MST model fails to generate 
the antonym \textit{Verkleinerung}.
This suggests that
target side constraints that are unseen during training may be ignored by the model, and that the models are not learning to systematically copy arbitrary constraints.

\subsection{Data Augmentation Experiment}

The results of our qualitative exploration suggests that the model 
would benefit from seeing more semantically divergent 
terminology constraints during training. To that end,
we propose a data augmentation experiment to increase the robustness of the APE models. We create novel training instances
by replacing the target language term constraint with
either a synonym or antonym (using Wiktionary), as well as
replacing it's occurrence in the post-edited target translation.
This results in 837,127 additional samples for eSCAPE corpus, and 2587 additional samples for official NMT data.
See \autoref{example_data_augmentation} for examples.

We then train MST append and MS LevT with the augmented pretraining corpus. We experiment with using augmented data only for pretraining (\textit{pretrain}) and for the fine-tuning process (\textit{ft}). The result can be seen in 
\autoref{result_augmentation}.

Interestingly, on the WMT19 test set, the data augmentation helps with TER and BLEU while having only a slight effect on Term\%. %
For the LevT, data augmentation helps all metrics.

Since the WMT19 APE test data contains few 
unusual constraints, the effect of the augmented data is relatively small. When we create antonym and synonyms examples from the WMT19 APE test data, we see fairly positive
trends, with pretraining and fine-tuning yielding additive
reductions in TER and gains in BLEU. This suggests that the augmentation method has a positive effect on 
the systematic copying behavior of the model.

\subsection{Post-Edit Stability}

To quantify the stability of the APE models, we compare the constrained APE output when given a target side synonym or antonym to the output of that 
same model under the original test set constraint using TER and BLEU.
Under this setting, higher BLEU and lower TER indicate that 
the model makes minimal changes when inserting a semantically 
divergent constraint. We also  report Term\%
to show how often the terminology was correctly translated given the input.
We refer to a model with high Term\% and BLEU but low TER as a stable model. 
Results of this experiment are 
shown in \autoref{result_syn_ant}. %

There are several takeaways from this 
experiment. First, the LevT TER scores are higher on average 
than the MST model suggesting that the 
LevT model is less stable, producing
different translations for each target
side constraint change. 

Second, as sensitivity to constraints increases (i.e. Term\% goes up), TER generally goes up, implying that 
models make more structural 
changes to the overall output in order to
accommodate constraints. Future work
on refinement tasks
like APE may benefit from including an explicit objective function to encourage output stability.

Finally,  synonyms are harder to translate than antonyms (i.e. Synonym Term\% is lower than Antonym Term\% for all models/training configurations). This may be because the original target side constraints are better represented in the 
decoder language model and are likely have higher probability than a synonym when
either could be plausibly used in the same
context. Antonyms may be less likely 
and therefore easier to override 
the preferences of decoder.

\section{Conclusion and Future Work}

This work introduces the terminology constrained APE task and several MST and LevT model variants for incorporating lexical constraints during 
post-editing. Furthermore, we show that constrained APE
is necessary for preserving lexical constraints in a MT to APE pipeline. 
Evaluations on standard APE benchmarks show that terminology constraints
are satisfied while improving the original MT quality. Finally, we show that the constrained APE models do not learn a robust systematic copying behavior, and propose a data augmentation method to help mitigate this issue.
In future work, we hope to explore ways of modifying model architecture or training algorithms to further improve the systematic copying behavior.

\section*{Acknowledgments}

This research is based upon work supported in
part by the Office of the Director of National Intelligence (ODNI), Intelligence Advanced Research
Projects Activity (IARPA), via contract \#FA8650-17-C-9117. The views and conclusions contained
herein are those of the authors and should not be
interpreted as necessarily representing the official
policies, either expressed or implied, of ODNI,
IARPA, or the U.S. Government. The U.S. Government is authorized to reproduce and distribute
reprints for governmental purposes notwithstanding any copyright annotation therein.

\clearpage

\bibliographystyle{acl_natbib}
\bibliography{main}

\clearpage

\clearpage

\end{document}